# Multi-Granular Attention-Driven Reinforcement Learning Framework for Web Intelligent Enhancement Systems

1st Navin Chhibber
*Infinity Tech Group*
Sunnyvale, CA, USA
naveenchibber.research@gmail.com

2nd Deepak Singh
*Gainwell Technologies*
USA
deepaksingh1981@gmail.com

3rd Anokh Kishore
*Capital One*
USA
anokhkishore@gmail.com

4th Nikita Chawla
*Independent Researcher*
USA
nikitachawla83@gmail.com

5th K. Anguraj
*Department of Electronics and Communication Engineering*
*Sona College of Technology*
Salem, India
k.anguraj@gmail.com

***Abstract*—From the past few years, web intelligent enhancement systems increasingly rely on heterogeneous and dynamic web data to deliver personalized, context-aware services. However, traditional machine learning, deep learning, and reinforcement learning models often struggle with semantic understanding, adaptability, and scalability in continuously evolving web environments. In this research, a Multi-Granular Attention-based Reinforcement Web Intelligent Enhancement System (MGAR-WIES) is proposed to address the challenges by integrating semantic graph modeling, attention mechanisms, and adaptive reinforcement learning. Initially, heterogeneous web data comprising structured, semi-structured and unstructured sources are collected and preprocessed for generating unified feature representations. These representations are transformed into a dynamic semantic graph, where entities and their relationships are modeled by using graph embeddings enhanced by attention mechanisms for capturing both local relevance and global contextual dependencies. Subsequently, an adaptive multi-agent reinforcement learning strategy leverages the attention-aware semantic states to optimize personalized web actions like content recommendation, navigation optimization, and service adaptation. Finally, the continuous online feedback is further integrated to update graph representations and learning policies in real time by ensuring sustained adaptability and performance. The proposed MGAR-WIES acheived better results in terms of accuracy (80%) when compared with existing approaches.**

## I. Introduction

The rapid growth of World Wide Web (WWW) has transformed it into massive, dynamic repository of heterogeneous data by encompassing text, images, videos, social interactions and real-time streams [1]. Further, this web intelligence enhancement systems have emerged as an important research area which aimed at extracting meaningful knowledge, patterns and insights from this complex web ecosystem [2]. These systems integrate artificial intelligence, data mining, machine learning, natural language processing and semantic web technologies for improving information retrieval, personalization, recommendation and decision support [3]. As web applications influenced the domains more like e-commerce, healthcare, education, smart cities and social networks, demand for intelligent systems which are capable for understanding user behavior and contextual information has deepened [4]. Moreover, web intelligence enhancement focuses not only on processing large-scale data but also on improving adaptability, accuracy and responsiveness of web-based services [5]. By enabling systems for learning from user interactions and evolving data, web intelligence plays an important role in delivering smarter as well as more user-centric web experiences [6]. Despite significant advancements, developing effective intelligence enhancement systems presents several challenges like sheer volume, velocity and variety of web data which makes real-time processing and analysis computationally demanding [7]. Simultaneously, the web data is unstructured, noisy, incomplete and dynamically changing by complicating accurate knowledge extraction. Similarly, the privacy and security concerns also pose serious challenges, as intelligent systems frequently depend on sensitive user data for delivering personalized services [8]. These systems ensure ethical data usage by maintaining system effectiveness which remained a complex task. Additionally, scalability and interoperability across various platforms and technologies are difficult for achieving [9]. The presence of biased data and algorithmic transparency issues affected the reliability of intelligent web systems [10]. Moreover, the adaption of intelligence models for evolving user preferences and emerging web trends required continuous learning mechanisms which are resource-intensive and technically challenging for implementing. The motivation for enhancing this web intelligence systems stems from growing importance for smarter, more adaptive and context-aware web applications [11]. As users increasingly expect personalized, accurate and timely information, traditional web systems faced more challenges in handling complex user requirements and massive data streams. This has enhanced the intelligence which improves decision-making, user satisfaction and operational efficiency among multiple domains [12]. In areas such as smart cities, healthcare analytics, online education and business intelligence, intelligent systems enabled the proactive responses, predictive insight and optimized resource utilization [13]. Furthermore, advancements in artificial intelligence and computational power provide new opportunities for overcoming the existing limitations and design robust enhancement frameworks [14]. By addressing challenges which are related to scalability, privacy and adaptability, web intelligence enhancement systems remove the gap between raw web data and actionable knowledge which drives innovation and supports development of next-generation intelligent web services [15]. The main contributions of this research include:

- A graph-based semantic knowledge modeling framework with attention mechanisms is developed for capturing the complex relationships among web entities. By constructing dynamic semantic graphs and applying attention-driven propagation, framework effectively preserved both local relevance and global contextual dependencies by improving semantic understanding and interpretability.
- An adaptive multi-agent reinforcement learning strategy is introduced for optimizing personalized web actions like content recommendation, navigation optimization and service adaptation. The use of semantically enriched graph representations as state inputs reduces state-space complexity, accelerates policy convergence and enhances decision stability under dynamic web environments.
- A continuous online adaptation and feedback integration module is incorporated for enabling real-time updating of semantic embeddings and learning policies. By developing the user interaction feedback and streaming data, the proposed approach maintains long-term personalization accuracy and robustness, effectively addressing concept drift and evolving user preferences.

This research is organized as follows: Section 2 reviews related works, Section 3 presents the proposed MGAR-WIES methodology in detail, Section 4 evaluates the performance of proposed framework, Section 5 discusses the experimental results. Finally, Section 6 concludes paper and outlines future research directions

## II. Literature Review

Gao et al. [16] introduced the Com-DDPG model for task offloading in Internet of Vehicles (IoV) environments using multiagent reinforcement learning. The model coordinated multiple agents to learn task dependencies, resource consumption and queue states with LSTM predicting internal states and BRNN enhancing inter-agent communication. While it improved the offloading performance and reduces latency compared to baseline methods, its limitations include high computational complexity and dependency on accurate environmental state predictions. However, the challenges presented in handling large-scale vehicular networks, dynamic mobility patterns and real-time adaptation under heterogeneous MEC infrastructures. Kadhim et al. [17] developed an automated system for online engineering education leveraging IoT and mobile wireless networks. Their framework used an Adaptive Layered Bayesian Belief Network (AL-BBN) for student activity classification, fuzzy logic for score calculation and Multi-Gradient Boosting Decision Tree (MGBDT) for decision-making. It effectively enhanced the virtual learning assessment over traditional manual methods, but approach suffered from sensor inaccuracies, high data preprocessing needs and limited scalability for large student cohorts. Still the challenges are involved in integrating diverse IoT devices seamlessly and ensuring real-time feedback for effective personalized online learning.

Kaouni et al. [18] designed an AI-based adaptive e-learning model to personalize hybrid online learning environments. The system dynamically adapted content based on learner preferences, intellectual abilities and learning patterns by providing tailored digital spaces through LMS platforms. While this method offered greater personalization than the traditional LMS, it struggled need for accurate learner profiling, computational overhead and potential biases which were presented in AI-driven recommendations. The main challenges were scaling system for large user bases by ensuring cross-platform compatibility and continuously updating AI models for reflecting the evolving learner behaviors.

Aminu et al. [19] suggested real-time threat intelligence and adaptive defense mechanisms for cyber threat detection. The framework integrated high-speed data processing (Apache Kafka, Spark Streaming), Machine Learning (ML) for dynamic defense techniques like Moving Target Defense and SDN. Although the model had improved proactive threat mitigation, system faced the challenges like integration complexity with legacy systems, workforce upskilling requirements and the challenges in visualizing actionable insights. However, the challenges were presented like optimizing real-time response, fine-tuning AI models for evolving threats and collaborative threat intelligence sharing across organizations.

Bharathi et al. [20] implemented an AI-driven adaptive learning architecture for business simulation games by adjusting learning paths and exercises based on student feedback and performance. By applying ML, it enhanced the skill retention, decision-making and creative problem-solving which were presented in corporate education contexts. While the model was effective in personalized learning, challenges still involved in high computational requirements, dependence on accurate learner profiling and gaps in evaluating the real-world business complexities. However, the challenges are scaling for diverse learner populations, integrating with existing curricula and continuously refining AI models for matching rapidly evolving business environments.

## III. Methodology

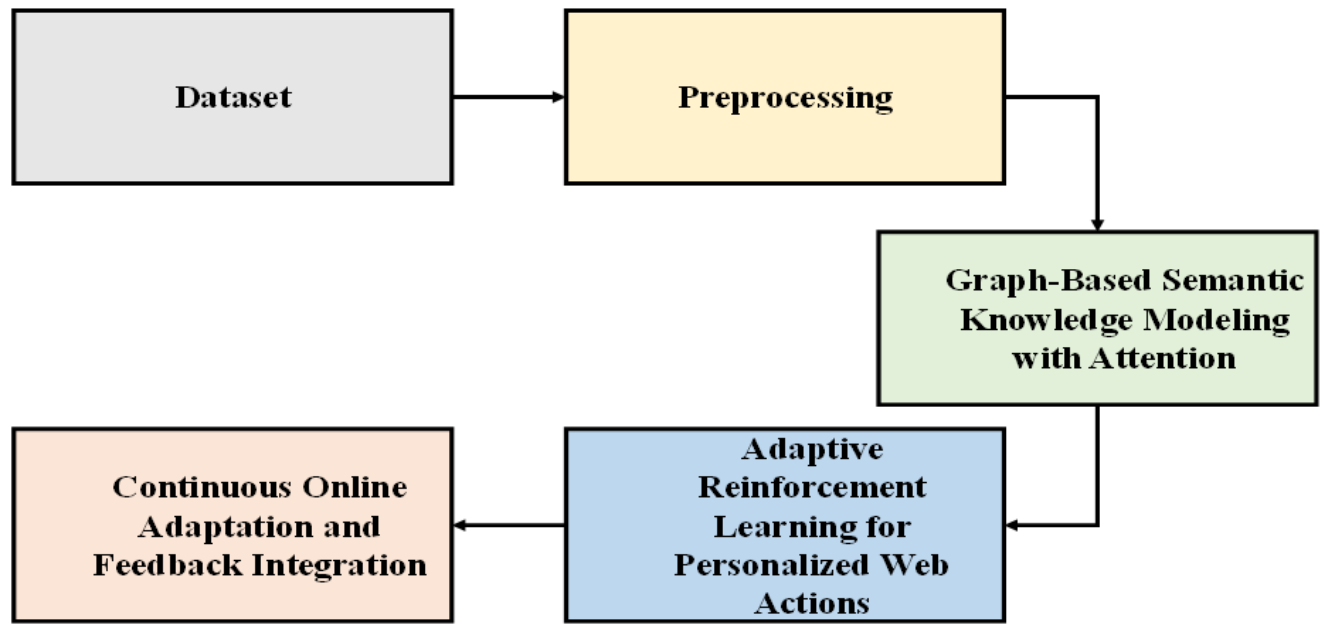

Fig. 1. Overview of proposed approach for web enhancement systems

The methodology of this research presents MGAR-WIES, an adaptive framework for web intelligent enhancement systems which integrates heterogeneous web data, semantic graph modeling, attention mechanisms and reinforcement learning. Initially, diverse web data are collected and preprocessed for generating the unified feature representations. These features are transformed into dynamic semantic graphs, where attention-based propagation captures contextual relationships among web entities. Then, adaptive multi-agent reinforcement learning strategy exploits these enriched representations for delivering the personalized web actions. After that, the continuous online feedback is incorporated to update both graph representations and learning policies in real time which ensures scalability, adaptability and sustained personalization performance in dynamic web environments. The overview of proposed methodology is shown in Figure 1.

### *A. Dataset*

The foundation of web intelligent enhancement system lies in collecting high-quality, heterogeneous dataset which reflects dynamic nature of web interactions. Modern web environments generate data in multiple modalities like textual content from blogs, articles and social media ($D_t$), visual data from images, infographics and video thumbnails ( $D_v$ ), behavioral data representing user interactions, clickstreams and session logs ($D_b$) and network data representing hyperlinks, social connections and also relational structures ( $D_n$ ). Formally, dataset is expressed as given in below Equation (1):

$$\mathcal{D}_f = \{D_t, D_v, D_b, D_n\} \quad (1)$$

The textual data is collected through web crawlers or APIs, while visual data extracted from page elements or multimedia repositories. Moreover, the behavioral logs track user navigation, dwell times and interaction sequences, also the network data captures structure of pages or social interactions in a graph-like form. Each data modality is collected with aim of representing both content and context by allowing multi-modal analysis and intelligent enhancement of web services. This dataset has challenges handling large-scale heterogeneous data by changing content and incomplete or missing information which require robust preprocessing and normalization strategies in subsequent stages.

### *B. Preprocessing*

Once multi-modal dataset $\mathcal{D}_f$ is acquired, preprocessing ensures its quality and suitability for intelligent modeling where this preprocessing encompasses cleaning, normalization, feature extraction and dimensionality reduction, each critical to maintaining integrity of multi-modal information like Noisy, duplicate, or missing entries are removed to generate a cleaned dataset as given in below Equation (2):

$$\mathcal{D}_f^c = \mathcal{D}_f \setminus \mathcal{N} \quad (2)$$

Where $\mathcal{N}$ represents inconsistencies or irrelevant data. Then, the normalization ensures uniform feature scales across modalities. For any feature $x_i \in X$, the normalized value as given in below Equation (3):

$$x_i' = \frac{x_i - min(X)}{max(X) - min(X)} \quad (3)$$

To reduce computational complexity while preserving critical information, the Principal Component Analysis (PCA) is applied as given in below Equation (4):

$$f_i' = P f_i, P \in \mathbb{R}^{k \times n}, k \ll n \quad (4)$$

The output of preprocessing is the fused, cleaned, normalized, and dimensionally reduced dataset as given in below Equation (5):

$$\mathcal{D}_f^p = \{f_1', f_2', \dots, f_m'\} \quad (5)$$

This dataset ensures that all subsequent modeling stages operate on coherent, high-quality, and computationally tractable representations while addressing heterogeneity, missing data and noise in multimodal web information.

### *C. Graph-Based Semantic Knowledge Modeling with Attention*

The preprocessed multi-modal dataset serves as basis for constructing semantic knowledge graph which encodes entities, their features and their relationships in structured form. Let $\mathcal{E} = \{\varepsilon_1, \varepsilon_2, \dots, \varepsilon_n\}$ denote entities and $\mathcal{R} = \{r_1, r_2, \dots, r_m\}$ represent relationships among them. The semantic graph is expressed as given in below Equation (6):

$$G = (\mathcal{E}, \mathcal{R}) \quad (6)$$

Each edge $\tau_k$ represents a connection between entities $\varrho_i$ and $\varrho_j$ , and the relationships are weighted by semantic similarity. A similarity function quantifies the closeness of entity features as given in below Equation (7):

$$S(e_i, e_j) = \frac{|F(e_i) \cap F(e_j)|}{|F(e_i) \cup F(e_j)|} \quad (7)$$

Where $F(e_i)$ denotes feature set of entity $e_i$. To encode relational dependencies, the entities are mapped into a low-dimensional embedding space $\mathrm{v}_i \in \mathbb{R}^d$ by using graph embedding techniques as given in below Equation (8):

$$v_i = Embed(G, e_i) \quad (8)$$

To enhance model's capability to prioritize important relationships, graph attention networks (GATs) are applied. The attention coefficient between node $i$ and neighbor $j$ is calculated as given in below Equation (9):

$$\alpha_{ij} = \frac{exp\left(LeakyReLU\left(a^T[Wv_i \| Wv_j]\right)\right)}{\sum_{k \in N_i} exp\left(LeakyReLU\left(a^T[Wv_i \| Wv_k]\right)\right)} \quad (9)$$

Where $W$ is a weight matrix, a represents learnable attention vector and $\mathcal{N}_i$ represents set of neighbors of node $i$. The updated node embeddings as given in below equation (10):

$$v_i' = \sigma\left(\sum_{j \in N_i} \alpha_{ij} W v_j\right) \quad (10)$$

This attention mechanism allows system to dynamically weigh relationships based on contextual relevance by improving content recommendation, link prediction and personalized service delivery. The embeddings $\mathrm{v}_i'$ preserve both structural and semantic information by making them suitable inputs for decision-making models in next stage. The challenges addressed in this stage are ontology incompleteness, ambiguity in entity representation and computational complexity for large-scale graphs. The use of attention enabled the prioritization of interactions by mitigating noise from irrelevant or weakly connected nodes. The graph structure also allows multi-modal integration, as textual, visual, behavioral and network features are all embedded within node representations by creating a unified semantic space for intelligent reasoning.

### D. *Adaptive Reinforcement Learning for Personalized Web Actions*

After the semantic graph embeddings are generated, next stage focuses on adaptive decision-making to enhance user experience on web platforms. By developing the low-dimensional node embeddings $v_i'$, system dynamically optimize content recommendations, navigation and personalization. Traditional static recommendation systems are limited because they cannot adapt to evolving user preferences or contextual changes. Therefore, Reinforcement Learning (RL) is introduced to model web system as an agent interacting with dynamic environment. The RL agent observes current state of system $s_t$ at time $t$ which comprises three components as given in below Equation (11):

$$s_t = \{v_t', U_t, C_t\} \tag{11}$$

Where $v_i'$ represents updated embedding of entity $i, U_t$ represents user behavioral context (e.g. clicks, scrolls, dwell times) and $C_t$ denotes content-related features of web page or resource. The agent selects action $a_t \in A$ like recommending page by prioritizing links or highlighting multimedia content based on a policy $\pi(a_t \mid s_t)$. The immediate reward $r_t$ reflects effectivness of action, quantified by metrics like user engagement, session duration or click-through rate. The cumulative reward over a session of length $T$ is expressed as given in below Equation (12):

$$R = \sum_{t=0}^{T} \gamma^t r_t \tag{12}$$

Where $\gamma \in [0,1]$ represents discount factor that balances importance of immediate versus future rewards. The RL agent's objective is to maximize $R$ by learning an optimal policy $\pi^*$ as given in below Equation (13):

$$\pi^* = arg\max_{\pi} E[R \mid \pi] \tag{13}$$

To handle complex, multi-faceted web environments, multi-agent reinforcement learning (MARL) framework is adopted. Here, multiple agents represent distinct web services-content, advertising. Navigation and social recommendations-that cooperate or compete to optimize global objective by enhancing overall user satisfaction. Agents communicate by shared embeddings $v_i'$ and coordinate through joint policies by combining individual service objectives with collective user engagement. The action-value function $Q(s_t, a_t)$ predicts expected rewards for taking action $a_t$ in state $s_t$ and is updated using temporal difference (TD) learning as given in below Equation (14):

$$Q(s_t, a_t) \leftarrow Q(s_t, a_t) + \alpha \left[r_t + \gamma \max_{a} Q(s_{t+1}, a) - Q(s_t, a_t)\right] \tag{14}$$

Where $\alpha$ represents learning rate which allows system to iteratively refine its strategy based on feedback by balancing exploration of new content with exploitation of known user preferences. By using RL in this stage is clear web environment is highly dynamic, users' preferences evolve over time and actions have long-term consequences on engagement. RL allows system to adapt in real time, optimize personalized actions for individual users and coordinate multi-agent objectives without requiring manual rules or static heuristics. Furthermore, by integrating semantic embeddings, system ensures that decisions are context-aware and grounded in relationships and content relevance encoded in graph. This stage overcomes the challenges include ensuring real-time inference, convergence of policies in large-scale multi-agent settings and avoiding overfitting to transient user behaviors.

### E. *Continuous Online Adaptation and Feedback Integration*

After implementing adaptive RL for personalized web actions, system maintain relevance and continuously improve as users interact with the platform and new content emerges. This focuses on continuous online adaptation and feedback integration which ensures that system evolves in real time to accommodate changing user behavior, trending content and dynamic web structures. This stage closes loop in proposed MGAR-WIES framework by transforming reactive model into fully self-improving, context-aware system. User feedback and engagement metrics like clicks, dwell time, scrolling patterns and content ratings which serve as immediate signals that are integrated into system. Let $F_u^t$ represent feedback data collected at time $t$ for user interactions, content quality and session dynamics and node embeddings $v_i'$ which are generated are updated iteratively for incorporating this new information as given in below Equation (15):

$$v_i^{(t+1)} = v_i^{(t)} + \eta \nabla_{v_i} \mathcal{L}(v_i; F_u^t) \tag{15}$$

Where $\eta$ represents learning rate and $\mathcal{L}$ represents loss function quantifying discrepancy between predicted outcomes and actual user responses. This embedding update mechanism allows semantic graph to adapt dynamically by reflecting evolving relationships among content, users and interactions. The RL policies from previous stage are also adjusted in real time based on cumulative feedback.

This ensures that the system does not become stable and refines strategies for maximizing the long-term user satisfaction. In multi-agent settings, joint policy updates synchronized the learning across agents by maintaining coordination among content recommendation, navigation optimization and advertisement placement. The online adaptation also incorporates novel content integration and cold-start mitigation. For newly added web pages or new users, initial embeddings are approximated via similarity with existing entities as given in below Equation (16):

$$v_{new} = \frac{1}{|\mathcal{N}_{sim}|} \sum_{j \subset \mathcal{N}_{sim}} v_j' \tag{16}$$

Where $\mathcal{N}_{\text{sim}}$ represents set of nearest similar entities based on feature similarity. This allowed immediate personalization without extensive historical data. Also, this stage effectively worked on continuous adaptation where the web environments and user preferences are highly dynamic and static models fail to sustain long-term engagement. Moreover, the real-time updates on the embeddings and policies enabled the context-aware, personalized and robust decision-making by improving the system's ability in responding to transient trends, seasonal behaviors or emerging topics. Additionally, by integrating feedback reduces errors because of its information, biases or incomplete knowledge graphs which ensured the higher accuracy and relevance.

## IV. Experimental Results

In this research, MGAR-WIES is proposed for enhancement systems and implemented on Python with libraries such as TensorFlow, PyTorch, NetworkX, and Scikit-learn, deployed on a workstation equipped with an Intel i7 processor, 32 GB RAM, NVIDIA GPU, and Ubuntu operating system. The performance of proposed MGAR-WIES is evaluated by using metrics like accuracy, precision and recall which are calculated as shown in below Equations (17) – (19):

$$Accuracy = \frac{TP+TN}{TP+TN+FP+FN} \times 100 \quad (17)$$

$$Precision = \frac{TP}{TP+FP} \times 100 \quad (18)$$

$$Recall = \frac{TP}{TP+FN} \times 100 \quad (19)$$

Here, $TP$ and $TN$ represents true positives and true negatives, $FP$ and $FN$ represents false positives and false negatives.

### A. Performance Analysis

The MGAR-WIES is proposed for cognitive radio networks with deep spectral sensing and compared with traditional approaches like Q-Learning, Deep RL and Deep Q-Networks (DQN) where the results are tabulated in below given Table 1.

TABLE I. Performance analysis of proposed MGAR-WIES with traditional models

| Performance Models | Accuracy (%) | Precision (%) | Recall (%) |
|---|---|---|---|
| Q-Learning | 75 | 74 | 72 |
| Deep RL | 76 | 75 | 73 |
| DQN | 78 | 77 | 74 |
| Proposed MGAR-WIES | 80 | 78 | 76 |

From the above Table 1, the proposed MGAR-WIES achieved better results in terms of accuracy (80%), precision (78%) and recall (76%) when compared with traditional approaches like Q-Learning, Deep RL and DQN respectively.

### B. Comparative Analysis

The proposed MGAR-WIES is compared with existing AL-BBN [17], those results are tabulated as shown in below Table 2.

TABLE II. Comparative Analysis Of Proposed MGAR-WIES With Existing Models

| Comparative Models | Accuracy (%) |
|---|---|
| AL-BBN [17] | 75 |
| Proposed MGAR-WIES | 80 |

From the above table 2, the proposed MGAR-WIES achieved better results in terms of accuracy (80%) when compared with existing AL-BBN [17] accuracy (75%) respectively.

## V. Discussion

The main objective of this research is to develop a model for web intelligent enhancement systems and the proposed MGAR-WIES effectively overcame the drawbacks of traditional and existing models like -Learning, Deep RL, DQN and AL-BBN [17] by its combination, multi-layered intelligence framework. The traditional approaches depend heavily on static feature representations and offline training which limited their ability to respond to dynamic web environments and evolving user behavior. The proposed MGAR-WIES addresses this limitation by incorporating continuous online learning and real-time feedback integration by enabling system for adapting knowledge representations and decision policies dynamically. The traditional models like Deep RL and DQN-based models improve adaptability but suffered from slow convergence, state-space explosion and limited semantic understanding of web content. Whereas, the proposed MGAR-WIES mitigated these issues by embedding graph-based semantic knowledge modeling with attention mechanisms which reduces state complexity by preserving contextual relationships among users, content and interactions. Unlike DQN which depends on discrete action spaces and value approximation alone, the proposed MGAR-WIES employs adaptive multi-agent RL guided by semantically enriched states by resulting in more stable learning and faster policy convergence. The existing AL-BBN [17] and similar probabilistic models primarily focused on uncertainty handling and classification but lack scalability and deep personalization capabilities. But, the proposed MGAR-WIES overcomes these constraints by combining probabilistic reasoning with graph embeddings and RL by enabling both uncertainty modeling and long-term decision optimization. Furthermore, attention-driven semantic graph enhances interpretability and relevance which ensured that critical web entities received higher priority during inference. Overall, the proposed MGAR-WIES delivers superior adaptability, scalability and personalization by making it more effective for complex, large-scale web intelligent enhancement systems when compared with existing approaches.

## VI. Conclusion

This research developed the effectiveness of multi-source web data integration for intelligent enhancement systems. Accordingly, adaptive and scalable the proposed MGAR-WIES framework is developed by integrating graph-based semantic modeling with attention mechanisms and adaptive RL. The proposed framework employs advanced feature learning techniques for encoding heterogeneous web content and user interaction data by generating modality-aware representations which are aligned within a unified semantic space for capturing contextual and behavioral patterns. Then, inter-entity relationships among users, content and services which are modeled by dynamic graph construction and attention-driven propagation by preserving both local relevance and global structural dependencies. After that, the personalized web actions are derived by using adaptive multi-agent RL, optimizing user engagement and system responsiveness under dynamic environments. The proposed MGAR-WIES acheived better results in terms of accuracy (80%) when compared with existing approaches. In the future, this work will focus on integrating large-scale streaming data, multimodal social and sensory signals also enhancing model transparency through explainable AI techniques to support trustworthy and sustainable deployment.